\documentclass{article}
\usepackage{url}
\usepackage[pdftex]{graphicx}
\usepackage{amsmath}
\usepackage{booktabs}
\usepackage{amsfonts}
\usepackage{algorithm}
\usepackage{algpseudocode} 
\usepackage{float}
\usepackage{multirow}
\usepackage{tcolorbox}
\usepackage{authblk}  % Added for better author/affiliation formatting
\usepackage{booktabs}      % For professional table rules
\usepackage{adjustbox}     % For resizing tables
\usepackage{caption}       % For better caption placement
\usepackage{tabularx}      % For dynamic column widths
\usepackage{natbib} 

\title{Towards Label-Free Biological Reasoning Synthetic Dataset Creation via Uncertainty Filtering}
\author[1]{Josefa Lia Stoisser}
\author[1]{Lawrence Phillips}
\author[1]{Aditya Misra}
\author[2]{Tom Lamb}
\author[2]{Philip Torr}
\author[1]{Marc Boubnovski Martell}
\author[1]{Julien Fauqueur}
\author[1]{Kaspar Märtens}

\affil[1]{Novo Nordisk}
\affil[2]{University of Oxford}

\date{October 2025} % Optional: update the date as needed

\begin{document}

\maketitle

\begin{abstract}
Synthetic chain-of-thought (CoT) traces are widely used to train large reasoning models (LRMs), improving generalization by providing step-level supervision. Yet most approaches require ground-truth labels to seed or filter these traces—an expensive bottleneck in domains like biology where wet-lab data are scarce.
We propose a label-free alternative: \emph{uncertainty-based filtering}, which uses a model’s {own confidence}—quantified through established uncertainty metrics like self-consistency and predictive perplexity—as a substitute for external labels. We sample multiple reasoning traces and retain only low-uncertainty subsets. Applied to biological perturbation prediction, a domain where wet-lab labels are especially costly, we show that the filtered subset has higher accuracy, and that supervised fine-tuning (SFT) on uncertainty-filtered data outperforms unfiltered synthetic data, narrows the gap to ground-truth training, and surpasses strong LRM baselines. Ablations show that per-class filtering corrects for class-specific uncertainty scales and that hybrid uncertainty metrics yield higher-quality datasets. Our results suggest that model-internal confidence is a powerful signal for efficient reasoning dataset creation, enabling LRMs in domains where supervision is expensive.
\end{abstract}

%%%%%%%%%%%%%%%%%%%%%%%%%%%%%
\section{Introduction}

Synthetic chain-of-thought (CoT) traces have become a cornerstone for training large reasoning models (LRMs), providing step-level supervision that improves generalization across mathematics, coding, and symbolic tasks \citep{guo2025deepseek, li2023symbolic, guha2025openthoughts}. However, most pipelines for generating such traces rely on ground-truth labels to filter sampled generations \citep{guo2025deepseek, li2023symbolic, guha2025openthoughts}. While feasible in domains with abundant labels or automatic checkers, this creates a bottleneck where high-quality labels are costly or unavailable.

Applications in biology particularly highlight this challenge. Ground-truth labels, when available at all, often require costly experimental measurement, limiting the scale of supervision. In particular, \emph{cellular perturbation prediction}—predicting how a given perturbation (e.g. drug or gene knockout) affects target gene expression levels (up, down, or unchanged)—is a fundamental task underlying drug discovery and disease modeling. The challenge is compounded by fundamental epistemic uncertainty: even when outcomes can be measured, the underlying causal mechanisms (e.g. gene regulatory networks) remain poorly understood, precluding external validation of synthetic reasoning traces \citep{tejada2025causal}. Moreover, approaches that distill carefully curated reasoning traces into open-source models have shown to achieve task-specific performance that exceeds that of frontier LRMs \citep{synthpert}.

We address these challenges with \emph{uncertainty-filtered synthetic reasoning}. Our method samples multiple reasoning traces per example and filters them using the model’s \emph{own confidence}—quantified by self-consistency and predictive perplexity—without any external supervision or verifiers. This approach aims to simultaneously mitigate label scarcity (by reducing dependence on wet-lab outcomes) and guard against epistemic gaps (by discarding examples where the model itself is least confident), to yield cleaner and more reliable synthetic training data. Our contributions are threefold:
\begin{itemize}
\item We introduce a \emph{label-free dataset curation pipeline} that filters synthetic reasoning traces by uncertainty, enabling efficient reasoning data construction in unlabeled domains.
\item Applied to \emph{biological perturbation prediction} and evaluated on the established {PerturbQA} benchmark, we show that filtering traces by internal uncertainty yields subsets with higher accuracy on final predictions. Moreover, training on uncertainty-filtered data outperforms unfiltered synthetic data and narrows the gap to ground-truth training, reducing reliance on costly wet-lab experiments.

\item Through ablations, we find that per-class filtering corrects for class-specific uncertainty scales and that using an existing hybrid uncertainty score \cite{Vashurin2025} leads to highest quality synthetic data, suggesting \emph{general principles for data-efficient LLM reasoning}.
\end{itemize}

%%%%%%%%%%%%%%%%%%%%%%%%%%%%%
\section{Background}

\paragraph{Synthetic Reasoning Datasets.}
Chain-of-thought (CoT) prompting~\citep{wei2022cot} elicits intermediate reasoning steps, and reliability can be improved via sampling and consistency filtering~\citep{zhang2022autocot, wang2022selfconsistency}. LLM prompting has become a general tool for generating synthetic datasets~\citep{zhu2024deepseekprover, goyal2024systematic, wang2022self}. Recent work combines these directions to produce synthetic reasoning traces~\citep{guo2025deepseek, li2023symbolic, guha2025openthoughts, zelikman2022star, yu2025cot, stoisser2025struct, stoisser2025sparks}. However, most methods still require ground-truth labels~\citep{guo2025deepseek, li2023symbolic}, sometimes falling back on heuristics like self-consistency when labels are absent~\citep{zelikman2022star, yu2025cot}, and are therefore not fully label-free.

\paragraph{Biological Perturbation Prediction.}
Predicting how genetic or chemical perturbations alter cellular states is a central challenge in computational biology. Many aspects of perturbation prediction remain unknown \citep{tejada2025causal} , making it a canonical testbed for label-scarce domains. Classical approaches (e.g., GEARS~\citep{roohani2024predicting}, scGPT~\citep{cui2024scgpt}) often underperform simple baselines~\citep{ahlmann2024deep, kernfeld2023systematic}. Recent LLM-based models—GenePT~\citep{chen2024genept}, SUMMER~\citep{wu2025contextualizing}, and SynthPert~\citep{synthpert}—adapt embeddings, retrieval, or synthetic reasoning. SynthPert shows that CoT traces can aid generalization, but the method relies on labeled outcomes. 

\paragraph{LLM Uncertainty Quantification.}
LLM uncertainty measures~\citep{shorinwa2025survey, geng2023survey, stoisser2025towards} include predictive entropy and probability margins~\citep{xiao2021hallucination, ling2024uncertainty}, or consistency-based agreement across multiple generations~\citep{abbasi2024believe, chen2023quantifying}. Hybrid methods such as CoCoA show that combining both yields the strongest correlation with correctness~\citep{Vashurin2025}, which we use as a substitute for lables.

%%%%%%%%%%%%%%%%%%%%%%%%%%%%%%%%%
\section{Methods}

\paragraph{Problem formulation.}  
We study perturbation prediction: given a tuple $(c, g, p)$ of cell type $c$, perturbation $p$, and gene $g$, the task is to predict whether $g$'s expression {increases} (\texttt{up}), {decreases} (\texttt{down}), or remains {unchanged} (\texttt{non-} \texttt{regulated}). This three-class formulation follows \cite{wu2025contextualizing} and reflects realistic biological workflows. As prior knowledge of gene responses is often unavailable, we generated synthetic reasoning data for this task, as illustrated in Figure \ref{fig:eff_fic}.

{
\setlength{\textfloatsep}{5pt}
\begin{figure}[t]
  \centering
  \includegraphics[width=\linewidth]{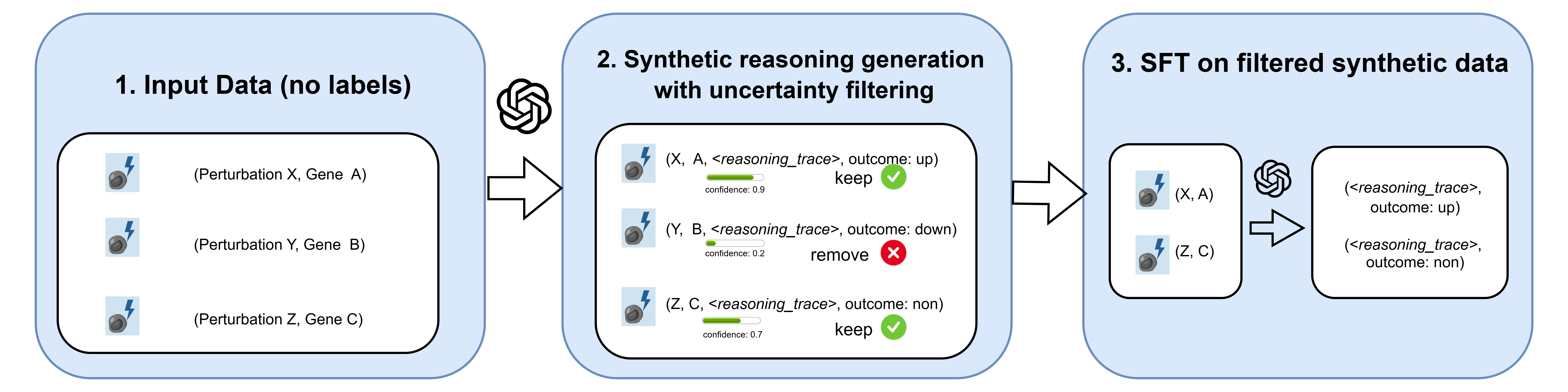}
    \caption{\textbf{Uncertainty-filtered synthetic reasoning pipeline.}
Step 1: Generate multiple synthetic reasoning traces with predicted outcomes from unlabeled perturbation–gene pairs.
Step 2: Score each trace for internal uncertainty (self-consistency and perplexity) and retain only low-uncertainty traces.
Step 3: Use the retained traces as a label-free dataset for supervised fine-tuning (SFT), improving reasoning models without ground-truth labels.}
  \label{fig:eff_fic}
\end{figure}
}
\paragraph{Synthetic reasoning generation.}  
To address the scarcity of labeled perturbation outcomes, we generate synthetic chain-of-thought (CoT) traces using a frontier LLM. For each input tuple $(c,g,p)$, we produce $k+1$ reasoning paths: one greedy-decoded trace $r_0$ and $k$ high-temperature sampled traces $\{r_i\}_{i=1}^k$, keeping $r_0$ as the response. Each trace consists of a natural-language explanation paired with a final prediction in $\{\texttt{up}, \texttt{down}, \texttt{non-} \texttt{regulated}\}$. See appendix \ref{app:efficiency} for a runtime analysis.

\paragraph{Uncertainty filtering.}
We estimate the reliability of synthetic traces using the \emph{CoCoA} metric~\citep{Vashurin2025}, which combines semantic consistency across sampled traces $\{r_i\}_{i=0}^k$ with predictive perplexity of $r_0$.  For each class, we retain the top-$x\%$ traces with the lowest CoCoA, ensuring balanced coverage across outcomes. Details are in appendix \ref{app:cocoa}.

\paragraph{Supervised fine-tuning.}  
Finally, we perform supervised fine-tuning (SFT) of a base LLM on the filtered dataset. The model is trained to reproduce both the reasoning trace and the final prediction, distilling patterns from the synthetic pool. We choose SFT over reinforcement learning objectives, since in this setting the base model is not strong enough for a cold-start RL setup. This procedure does not require ground-truth experimental labels, enabling adaptation to perturbation prediction under extreme label scarcity.

%%%%%%%%%%%%%%%%%%%%%%%%%%%%%%%%%
\section{Experiments}

\paragraph{Dataset.}
We evaluate on the established \textit{PerturbQA} benchmark \citep{wu2025contextualizing}, which reformulates perturbation prediction as natural language tuples (cell type, perturbation, gene) with labels {\texttt{up}, \texttt{down}, \texttt{non-} \texttt{regulated}}. Class imbalance (\texttt{non-} \texttt{regulated} dominates) motivates per-class filtering. We generate 48k synthetic traces and retain the top $x=10$ percent per class under CoCoA \citep{Vashurin2025}, yielding a training set of  4.8k samples. We use the official train/test split provided with the dataset.

\paragraph{Baselines.}
We benchmark our label-free approach against both zero-shot and supervised baselines:
(i) \textit{Zero-shot}: out-of-the-box performance of teacher and student models.
(ii) \textit{Ground Truth + Synthetic Data SFT \citep{synthpert}}: Augmentation with label-conditioned synthetic traces, followed by filtering to retain only those with correct predictions. This represents the best-performing label-dependent strategy.
(iii) \textit{Random-sampling (10\%)}: a size-matched control that selects traces uniformly at random, isolating the effect of uncertainty filtering.
(iv) \textit{Unfiltered (100\%)}: training on the entire synthetic pool, testing whether more data alone suffices.

We use Gemini 2.5 Pro for synthetic data generation due to strong reasoning performance and access to token-level log-probabilities, with results for Qwen3-32B model in Appendix \ref{app:qwen}, and we train Qwen3-32B. Implementation details are reported in Appendix \ref{app:training}. We report means with stratified bootstrapped standard errors (5,000 resamples) for each metric.

%%%%%%%%%%%%%%%%%%%%%%%%%%%%%%%%%
\subsection{Main Results}

\paragraph{Lower-uncertainty subsets contain more predictive traces.}
%\textbf{Uncertainty as a Proxy for Synthetic Data Quality.}
%Table~\ref{tab:synth_data} shows that subsets with lower CoCoA scores contain a higher fraction of correct answers. Accuracy and F1 rise steadily from the $100\%$ to $1\%$ lowest-uncertainty subsets, suggesting that lower uncertainty subsets have higher quality. Appendix Figures~\ref{plot}–\ref{plot1} further illustrate this trend: bar plots over decile splits show that groups with lower uncertainty consistently achieve higher F1 scores.
% Table\ref{tab:synth_data} shows that lower CoCoA scores concentrate correct answers: Acc rises from 0.42 on the full set (48k) to 0.49 on the most confident 1\% (480). Per-class F1 improves from 0.12 (Up) and 0.14 (Down) to 0.30 (Up) and 0.21 (Down), suggesting that lower uncertainty subsets have higher quality. Appendix Figures\ref{plot}–\ref{plot1} visualize this pattern across deciles, where lower-uncertainty groups consistently achieve higher F1. 
Table~\ref{tab:synth_data} shows that uncertainty filtering provides clear signal: quality improves monotonically as we retain progressively lower-uncertainty data. Accuracy rises from 0.42 (full 48k) to 0.49 (top 1\%, 480 examples), with consistent gains at every threshold. Per-class F1 for minority classes improves substantially—from 0.12/0.14 (Up/Down) to 0.30/0.21—demonstrating that uncertainty identifies not just correct predictions but more balanced, higher-quality reasoning. Appendix Figures~\ref{plot}–\ref{plot1} further visualize this trend across deciles.
As a qualitative check, expert annotation of sample traces confirmed this pattern: low-uncertainty examples contained sound biological reasoning, while high-uncertainty ones exhibited errors undermining their conclusions (Appendix~\ref{app:bio-annotation}).
% As an illustrative check, a biologist annotated two traces: the low-uncertainty example was correct, whereas the high-uncertainty one contained errors that undermined its conclusion (Appendix~\ref{app:bio-annotation}).

\paragraph{Finetuning on uncertainty filtered traces leads to better performance.}
Table~\ref{tab:main-results} shows that zero-shot Qwen3-32B achieves only $0.40$ accuracy. Ground-truth SFT remains strongest at $0.62$ accuracy. Among label-free methods, unfiltered training ($100\%$) reaches $0.52$, and random $10\%$ sampling falls to $0.48$. Crucially, our \emph{uncertainty-filtered} $10\%$ subset achieves $0.57$ accuracy and F1 scores of $0.26$ (Up) and $0.28$ (Down), substantially improving over both random and unfiltered subsets, and surpassing the strong LRM Gemini 2.5 Pro. This demonstrates that how data is selected matters: uncertainty filtering enables strong performance with only 10\% of synthetic traces, outperforming both random sampling at the same scale and unfiltered training on 10× more data.
%This shows that subset quality, not size, drives downstream gains. Despite reducing data volume, the filtered model retains generalization across genes, suggesting that increased data quality outweighs potential losses in data diversity.

\begin{table}[t]
\centering
\footnotesize
\begin{adjustbox}{max width=\textwidth}
\begin{tabular}{lccc|ccc|ccc|cc}
\toprule
\textbf{Data} & \multicolumn{3}{c}{\textbf{Up}} & \multicolumn{3}{c}{\textbf{Down}} & \multicolumn{3}{c}{\textbf{Non-reg.}} & \textbf{Acc} & \textbf{\# samples} \\
\midrule
& {Prec} & {Rec} & {F1} & {Prec} & {Rec} & {F1} & {Prec} & {Rec} & {F1} & & \\
\midrule
All data  & 0.07 $\pm$ 0.001 & 0.31 $\pm$ 0.005 & 0.12 $\pm$ 0.004 & 0.08 $\pm$ 0.003 & 0.64 $\pm$ 0.009 & 0.14 $\pm$ 0.003 & \textbf{0.93} $\pm$ 0.001 & 0.39 $\pm$ 0.001 & 0.55 $\pm$ 0.001 & 0.42 $\pm$ 0.009 & 48k \\
Top 20\% & 0.11 $\pm$ 0.006 & 0.34 $\pm$ 0.009 & 0.16 $\pm$ 0.011 & 0.10 $\pm$ 0.010 & 0.65 $\pm$ 0.012 & 0.17 $\pm$ 0.013 & 0.91 $\pm$ 0.010 & 0.55 $\pm$ 0.009 & 0.32 $\pm$ 0.009 & 0.44 $\pm$ 0.019 & 9.6k \\
Top 10\% & 0.12 $\pm$ 0.004 & 0.36 $\pm$ 0.006 & 0.18 $\pm$ 0.007 & 0.11 $\pm$ 0.008 & 0.66 $\pm$ 0.010 & 0.19 $\pm$ 0.009 & 0.88 $\pm$ 0.015 & 0.39 $\pm$ 0.007 & 0.54 $\pm$ 0.018 & 0.45 $\pm$ 0.013 & 4.8k \\
Top 5\% & 0.14 $\pm$ 0.002 & 0.35 $\pm$ 0.003 & 0.20 $\pm$ 0.005 & \textbf{0.12} $\pm$ 0.012 & 0.68 $\pm$ 0.014 & 0.20 $\pm$ 0.016 & 0.87 $\pm$ 0.021 & 0.39 $\pm$ 0.013 & 0.54 $\pm$ 0.024 & 0.46 $\pm$ 0.015 & 2.4k \\
Top 1\% & \textbf{0.23} $\pm$ 0.015 & \textbf{0.43} $\pm$ 0.020 & \textbf{0.30} $\pm$ 0.018 & \textbf{0.12} $\pm$ 0.017 & \textbf{0.69} $\pm$ 0.028 & \textbf{0.21} $\pm$ 0.020 & 0.88 $\pm$ 0.012 & \textbf{0.41} $\pm$ 0.018 & \textbf{0.56} $\pm$ 0.014 & \textbf{0.49} $\pm$ 0.023 & 480 \\
\bottomrule
\vspace{0.2 mm}
\end{tabular}
\end{adjustbox}
\caption{\footnotesize \textbf{Lower-uncertainty subsets contain more predictive traces.} on (Prec), recall (Rec), and F1 score of synthetic data generated zero-shot by Gemini 2.5 Pro via the approach described in Figure \ref{fig:eff_fic}. The rows correspond to the top $x\%$ of data retained after filtering by CoCoA metric. Lower uncertainty subsets achieve higher Acc and F1, indicating improved quality.}
\label{tab:synth_data}
\end{table}

{
\setlength{\textfloatsep}{5pt}
\begin{table}[t]
\centering
\footnotesize
\begin{adjustbox}{max width=\textwidth}\begin{tabular}{lccc|ccc|ccc|c}
\toprule
\textbf{Method} & \multicolumn{3}{c}{\textbf{Up}} & \multicolumn{3}{c}{\textbf{Down}} & \multicolumn{3}{c}{\textbf{Non-reg.}} & \textbf{Acc} \\
\midrule
 & Prec & Rec & F1 & Prec & Rec & F1 & Prec & Rec & F1 &  \\
\midrule
\multicolumn{11}{l}{\textbf{Zero-shot baseline}} \\
Zero-shot Gemini 2.5 Pro  & 0.20 ± 0.02 & 0.19 ± 0.01 & 0.20 ± 0.01 & 0.18 ± 0.01 & 0.22 ± 0.01 & 0.20 ± 0.01 & 0.79 ± 0.02 & 0.58 ± 0.02 & 0.67 ± 0.02 & 0.50 ± 0.02 \\
Zero-shot Qwen3-32B & 0.11 ± 0.01 & 0.27 ± 0.02 & 0.16 ± 0.01 & 0.18 ± 0.01 & 0.10 ± 0.01 & 0.13 ± 0.01 & 0.78 ± 0.02 & 0.45 ± 0.02 & 0.57 ± 0.02 & 0.40 ± 0.02 \\
\midrule
\multicolumn{11}{l}{\textbf{Label-based Training (Upper bound)}} \\
Ground truth + Synth data & 0.28 ± 0.02 & 0.51 ± 0.03 & 0.36 ± 0.02 & 0.16 ± 0.01 & 0.77 ± 0.02 & 0.27 ± 0.02 & 0.97 ± 0.01 & 0.61 ± 0.02 & 0.75 ± 0.02 & 0.62 ± 0.01 \\\midrule
\multicolumn{11}{l}{\textbf{Label-free Training}} \\
$100\%$-Unfiltered & 0.12 ± 0.04 & 0.31 ± 0.05 & 0.17 ± 0.04 & 0.17 ± 0.02 & 0.21 ± 0.03 & 0.19 ± 0.02 & 0.88 ± 0.02 & 0.59 ± 0.02 & 0.71 ± 0.02 & 0.52 ± 0.01 \\
$10\%$-Random-sampling & 0.11 ± 0.02 & 0.28 ± 0.03 & 0.16 ± 0.03 & 0.17 ± 0.02 & 0.19 ± 0.05 & 0.18 ± 0.03 & 0.79 ± 0.02 & 0.49 ± 0.03 & {0.60} ± 0.03 & 0.48 ± 0.03 \\
10\%-Uncertainty-filtered (Ours) & \textbf{0.22 ± 0.03} & \textbf{0.32} ± 0.04 & \textbf{0.26} ± 0.03 & \textbf{0.19} ± 0.02 & \textbf{0.55} ± 0.05 & \textbf{0.28} ± 0.02 & \textbf{0.91} ± 0.02 & \textbf{0.60} ± 0.01 & \textbf{0.72} ± 0.02 & \textbf{0.57} ± 0.02 \\
\bottomrule
\vspace{0.2 mm}
\end{tabular}
\end{adjustbox}
\caption{\footnotesize \textbf{Finetuning on uncertainty filtered traces leads to better performance.} Precision (Prec), recall (Rec), F1 per class (Up, Down, Non-regulated), and overall accuracy (Acc) are shown for zero-shot baselines (teacher: Gemini 2.5 Pro; student: Qwen3-32B), fully supervised methods (with/without synthetic reasoning traces), and label-free training on uncertainty-filtered, random, or full synthetic datasets. Best scores for label-free training are bolded.}
\label{tab:main-results}
\end{table}
}

\subsection{Ablation Studies}

\paragraph{Per-class filtering outperforms global selection.}
Table~\ref{tab:filtering-ablation} shows that random sampling keeps performance near unfiltered data (Up F1 $0.10$ vs $0.12$). Global filtering improves recall for Down ($0.71$ vs $0.64$), but collapses Non-regulated F1 to $0.12$, hurting overall accuracy ($0.16$). In contrast, per-class filtering maintains balance across classes (Up F1 $0.18$, Down F1 $0.19$, Non-reg. F1 $0.31$) and yields the highest accuracy ($0.25$), suggesting the importance of class-aware selection.

\paragraph{Hybrid uncertainty metrics outperform single signals.}
Table~\ref{tab:uncertainty-ablation} shows that perplexity alone performs worst (Up F1 $0.14$, Down F1 $0.15$). Consistency improves quality (Up F1 $0.16$, Down F1 $0.20$, Acc $0.24$). The hybrid CoCoA score achieves the strongest balance (Up F1 $0.18$, Down F1 $0.19$, Non-reg. F1 $0.31$, Acc $0.25$), suggesting that combining perplexity and self-consistency produces a more reliable label-free filter than either approach individually.

\section{Discussion}

We demonstrate that model-internal uncertainty enables label-free filtering of synthetic reasoning traces, addressing an underexplored axis of efficiency: supervision efficiency. While prior work focuses on algorithmic or system-level gains, experimental sciences and many other domains are constrained by the cost and availability of labels. On biological perturbation prediction—where each label requires costly wet-lab experiments—filtering by self-consistency and perplexity yields training data with higher accuracy than random or unfiltered baselines. Fine-tuning on this filtered data narrows the gap to fully supervised training without ground-truth labels. 

Our results suggest that LLMs can self-curate training data, decoupling reasoning improvements from costly supervision. Although validated on biology — a domain with both label scarcity and epistemic uncertainty about underlying mechanisms — our method is domain-agnostic and applicable to any setting where experimental data or supervision is costly.

\paragraph{Limitations and future directions.} Uncertainty filtering adds computational overhead for trace generation and scoring, though this cost is parallelizable and amortized across datasets (see Appendix \ref{app:efficiency}). Like any self-supervised signal, uncertainty filtering relies on the model's learned representations and may be sensitive to distribution shift. However, combining filtered synthetic data with even small amounts of ground-truth labels could provide validation and improve robustness. Future work should explore complementary uncertainty signals, validate generalization across diverse domains and shifts, and investigate such semi-supervised setups to improve robustness while reducing reliance on costly ground-truth labels.

\bibliographystyle{plain} % Choose the desired bibliography style
\bibliography{bib} % Ensure 'bib.bib' exists in the same directory as this LaTeX file

\appendix
\renewcommand{\thefigure}{A\arabic{figure}}
\renewcommand{\thetable}{A\arabic{table}}

%%%%%%%%%%%%%%%%%%%%%%%%%%%%%%%%%%%%%%%%%%%%%%%%%%%%%
\section{Uncertainty Definition} \label{app:cocoa}
The \emph{CoCoA} score \citep{Vashurin2025} combines semantic consistency and perplexity. Let $r_0$ be the greedy trace and $\{r_i\}_{i=1}^k$ the sampled traces. We define uncertainty as:
\begin{equation}
    \text{CoCoA}(x) = \frac{2}{k} \sum_{i=1}^{k} \big(1 - \text{sim}(r_0, r_i)\big) \cdot U_{\mathrm{PPL}}(r_0),
\end{equation}
where $\text{sim}(r_0, r_i)$ is the semantic similarity between $r_0$ and $r_i$ (computed via a cross-encoder \citep{liu2019roberta} as in \citep{Vashurin2025}), and $U_{\mathrm{PPL}}(r_0)$ is the perplexity of $r_0$. Higher CoCoA indicates higher uncertainty. For each class, we retain the top-$x\%$ traces with the lowest CoCoA, ensuring balanced coverage across outcomes.

Our aim is not to design a new uncertainty estimator — indeed, both perplexity and self-consistency are well-studied. Instead, our contribution is to demonstrate that when combined and repurposed, these familiar measures provide a practical, scalable criterion for filtering synthetic chain-of-thought traces, yielding label-free datasets that are significantly more effective for downstream fine-tuning.

%%%%%%%%%%%%%%%%%%%%%%%%
\section{Implementation details} \label{app:training}
Synthetic traces are sampled from Gemini 2.5 Pro. For each input we draw $k=8$ high-temperature (temperature=1, top-p=1.0, top-k=50) completions plus one greedy trace. The student is Qwen3-32B, fine-tuned with QLoRA ($1e^{-5}$ learning rate, batch size 4, 20 epochs) on one A100 GPU for 4 hours. We report per-class precision, recall, F1, and overall accuracy on the held-out test set, bootstrapped over 5,000 resampling iterations to compute standard errors and 95\% confidence intervals for each metric. Prompt are reported in \ref{app:prompts}.

We set $k=8$ samples per query because this provides a tractable balance between (i) sufficient diversity for self-consistency estimation and (ii) manageable computational cost. Larger k increases stability but with diminishing returns.”
We retain the top 10\% per class based on CoCoA because this threshold reliably yields a strong quality–quantity trade-off (see Table \ref{tab:synth_data}): retaining fewer examples leads to under-coverage of minority classes, while higher thresholds reintroduce noisy traces. In preliminary training sweeps (5–20\%), 10\% consistently provided the best downstream accuracy.

We plan to open source our code and dataset upon publication.

%%%%%%%%%%%%%%%%%%%%%%%%
\section{Visualising Uncertainty as a Dataset Quality Signal} 

\begin{figure}[t]
  \centering
  \includegraphics[width=\linewidth]{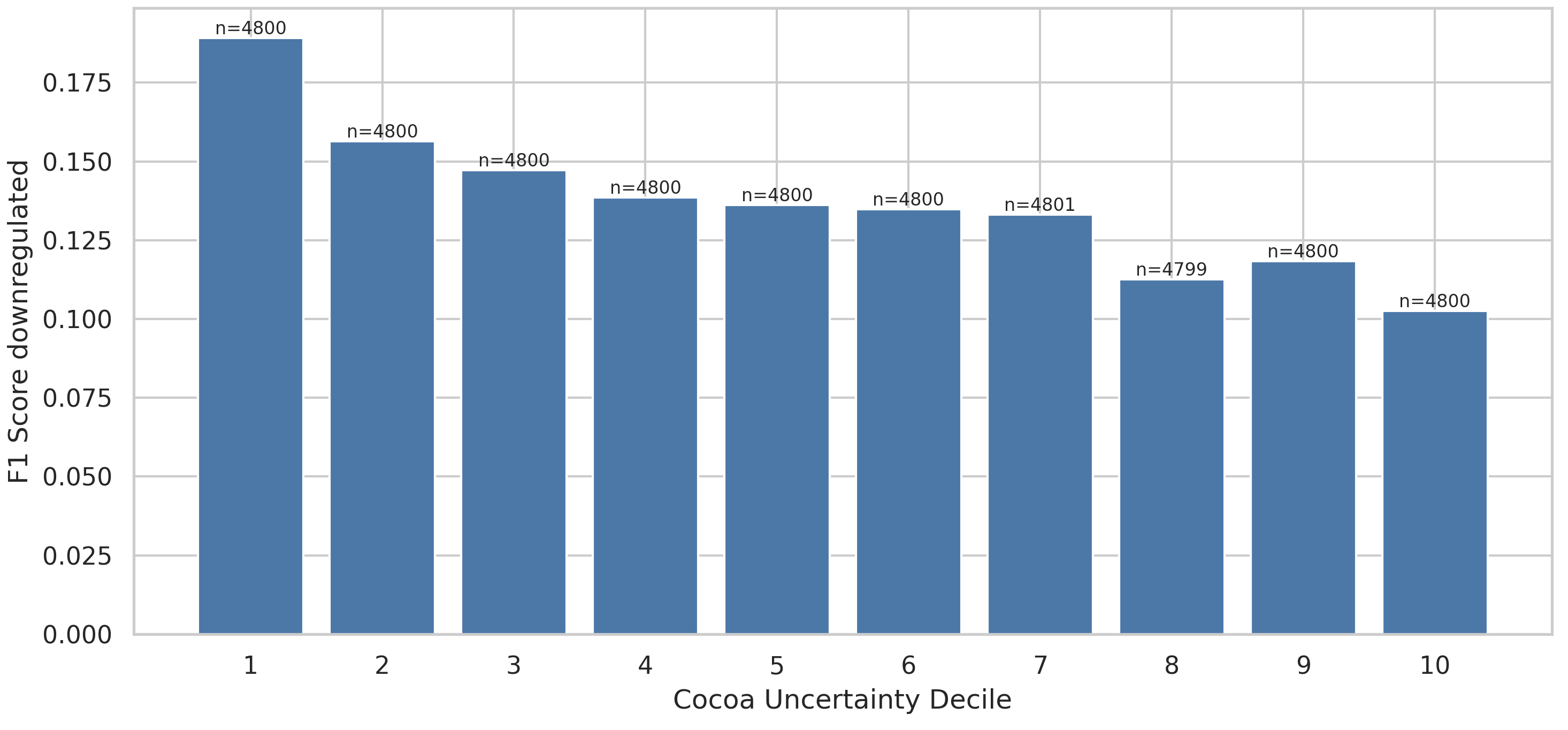}
    \caption{F1 score of upregulated genes stratified by CoCoA uncertainty deciles. Lower-uncertainty subsets yield consistently higher F1, with a clear monotonic trend across deciles. This confirms that uncertainty is strongly predictive of reasoning quality.}
  \label{plot}
\end{figure}

\begin{figure}[t]
  \centering
  \includegraphics[width=\linewidth]{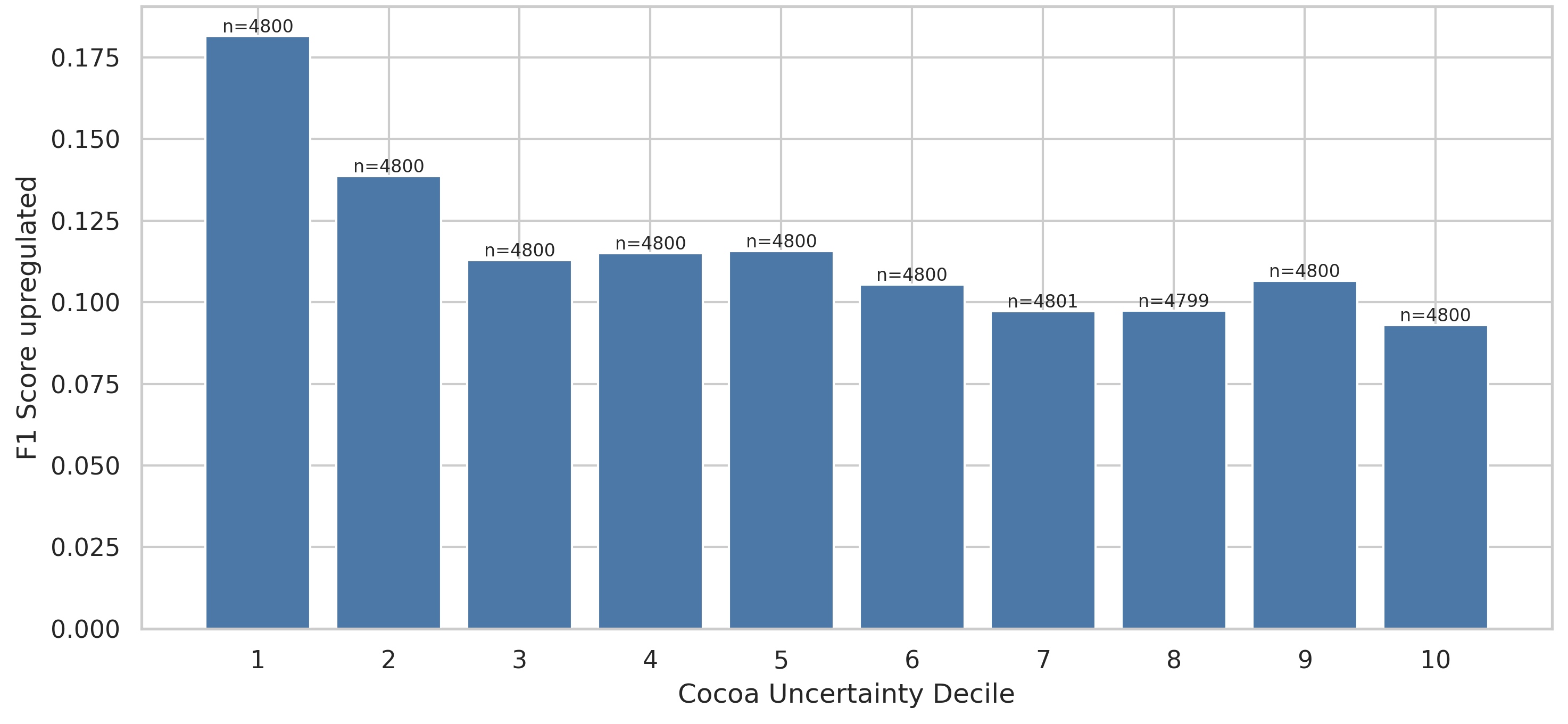}
    \caption{F1 score of downregulated genes stratified by CoCoA uncertainty deciles. Lower-uncertainty subsets yield consistently higher F1, with a clear monotonic trend across deciles. This confirms that uncertainty is strongly predictive of reasoning quality.}
  \label{plot1}
\end{figure}

We present visualizations of the fact that that groups with lower uncertainty consistently achieve higher F1 scores. Figures ~\ref{plot} \ref{plot1} show that F1 scores monotonically decrease across CoCoA uncertainty deciles, with the lowest-uncertainty bin yielding the cleanest traces.

% \begin{figure}[t]
%   \centering
%   \includegraphics[width=\linewidth]{Efficient_Reasoning/displot_example.png}
%     \caption{}
%   \label{plot}
% \end{figure}
%%%%%%%%%%%%%%%%%%%%%%%%

\section{Results for Qwen3-32B model} \label{app:qwen}
To illustrate the trends in uncertainty and data quality also for an open source model, we present a synthetic dataset generated by the Qwen3-32B model. The details are summarized in Table \ref{tab:synth_data_qwen33}.

\begin{table}[t]
\centering
\small
\begin{adjustbox}{max width=\textwidth}
\begin{tabular}{lccc|ccc|ccc|cc}
\toprule
\textbf{Synthetic Data} & \multicolumn{3}{c}{\textbf{Up}} & \multicolumn{3}{c}{\textbf{Down}} & \multicolumn{3}{c}{\textbf{Non-reg.}} & \textbf{Acc} & \textbf{\# samples} \\
\midrule
& {Prec} & {Rec} & {F1} & {Prec} & {Rec} & {F1} & {Prec} & {Rec} & {F1} & & \\
\midrule
1\% lowest uncertainty & \textbf{0.12} $\pm$ 0.025 & \textbf{0.33} $\pm$ 0.041 & \textbf{0.18} $\pm$ 0.033 & 0.08 $\pm$ 0.021 & \textbf{0.50} $\pm$ 0.052 & 0.14 $\pm$ 0.027 & \textbf{0.95} $\pm$ 0.018 & \textbf{0.54} $\pm$ 0.035 & \textbf{0.69} $\pm$ 0.030 & \textbf{0.53} $\pm$ 0.038 & 40 \\
5\% lowest uncertainty & \textbf{0.12} $\pm$ 0.015 & 0.27 $\pm$ 0.025 & 0.17 $\pm$ 0.020 & 0.09 $\pm$ 0.014 & 0.43 $\pm$ 0.032 & 0.14 $\pm$ 0.018 & 0.90 $\pm$ 0.010 & 0.54 $\pm$ 0.020 & 0.68 $\pm$ 0.017 & 0.51 $\pm$ 0.021 & 121 \\
10\% lowest uncertainty & 0.09 $\pm$ 0.012 & 0.23 $\pm$ 0.021 & 0.13 $\pm$ 0.016 & 0.07 $\pm$ 0.011 & 0.38 $\pm$ 0.025 & 0.11 $\pm$ 0.013 & 0.90 $\pm$ 0.009 & 0.53 $\pm$ 0.018 & 0.67 $\pm$ 0.015 & 0.50 $\pm$ 0.019 & 160 \\
20\% lowest uncertainty & 0.10 $\pm$ 0.010 & 0.29 $\pm$ 0.018 & 0.15 $\pm$ 0.013 & 0.07 $\pm$ 0.009 & 0.29 $\pm$ 0.021 & 0.11 $\pm$ 0.011 & 0.89 $\pm$ 0.008 & 0.53 $\pm$ 0.015 & 0.66 $\pm$ 0.013 & 0.50 $\pm$ 0.016 & 200 \\
All data (100\%) & 0.10 $\pm$ 0.003 & 0.24 $\pm$ 0.018 & 0.14 $\pm$ 0.014 & \textbf{0.10} $\pm$ 0.011 & 0.34 $\pm$ 0.032 & 0.15 $\pm$ 0.011 & 0.86 $\pm$ 0.004 & 0.53 $\pm$ 0.007 & 0.65 $\pm$ 0.007 & 0.49 $\pm$ 0.008 & 4000 \\

\bottomrule
\vspace{0.2 mm}
\end{tabular}
\end{adjustbox}
\caption{\textbf{Synthetic reasoning dataset quality using Qwen3-32B model.} Each row shows the precision (Prec), recall (Rec), and F1 score per class, as well as overall accuracy (Acc). Uncertainty filtering via CoCoA metric.}
\label{tab:synth_data_qwen33}
\end{table}

%%%%%%%%%%%%%%%%%%%%%%%%

\section{Illustrative Expert Annotation of Reasoning Traces} 
\label{app:bio-annotation}

We asked a PhD-trained biologist to annotate reasoning traces. Below is one drawn from the low-CoCoA (low uncertainty) subset and one from the high-CoCoA (high uncertainty) subset. The low-uncertainty trace was found to be correct throughout, whereas the high-uncertainty trace contained an early factual error that propagated and rendered the overall conclusion incorrect.

%We asked a PhD-trained biologist to annotate two example reasoning traces: one drawn from the low-CoCoA (low uncertainty) subset and one from the high-CoCoA (high uncertainty) subset.   

\subsection*{Low-Uncertainty Example (All Steps Correct)}

\begin{tcolorbox}[colback=blue!10!white, colframe=blue!75!black, title=Prompt]
Analyze the regulatory effect of knocking down the \texttt{ALG2} gene on the \texttt{PDIA6} gene in a single-cell K562 cell line using CRISPR interference.
\end{tcolorbox}

\begin{tcolorbox}[colback=blue!10!white, colframe=blue!75!black, title=Expert annotation (selected points):]
\begin{itemize}
    \item ALG2 functions in N-linked glycosylation (TRUE).  
    \item PDIA6 is an ER chaperone induced by UPR (TRUE).  
    \item ALG2 knockdown impairs glycosylation, induces ER stress, and activates UPR (TRUE).  
    \item UPR upregulates PDIA6 via XBP1s/ATF6 (TRUE).  
    \item Context: K562 cells are sensitive to ER stress (TRUE).  
\end{itemize}
\end{tcolorbox}

\textbf{Conclusion:} Expert judged the chain of reasoning correct, predicting \texttt{PDIA6} upregulation.  

\subsection*{High-Uncertainty Example (Early Factual Error)}

\begin{tcolorbox}[colback=blue!10!white, colframe=blue!75!black, title=Prompt]
Analyze the regulatory effect of knocking down the \texttt{CD3EAP} gene on the \texttt{RPTOR} gene in a single-cell K562 cell line using CRISPR interference.
\end{tcolorbox}

\begin{tcolorbox}[colback=blue!10!white, colframe=blue!75!black, title=Expert annotation (selected points):]
\begin{itemize}
    \item CD3EAP is incorrectly identified as calpastatin (FALSE).  
    \item Downstream reasoning (calpain hyperactivation → mTORC1 suppression) is partly biologically plausible, but depends on the incorrect gene assignment.  
    \item Final conclusion (RPTOR downregulation) judged not supported.  
\end{itemize}
\end{tcolorbox}

\textbf{Conclusion:} A single early factual error misdirects the reasoning chain, making the overall conclusion unreliable despite plausible intermediate statements.

%%%%%%%%%%%%%%%%%%%%%%%%

\section{Ablation Details}
Table~\ref{tab:filtering-ablation} evaluates different filtering strategies for the synthetic dataset. Random sampling selects traces uniformly, global uncertainty filtering selects the lowest 10\% CoCoA across all examples, and per-class filtering selects the lowest 10\% within each class. The results show that per-class filtering consistently yields higher-quality traces, as uncertainties are not directly comparable across classes; without per-class selection, minority classes tend to be underrepresented.

Table~\ref{tab:uncertainty-ablation} compares different uncertainty metrics for selecting synthetic traces: perplexity alone, consistency across multiple generations, and the CoCoA score, which combines both signals. The results indicate that using CoCoA produces the most reliable traces across all classes, demonstrating that combining perplexity and self-consistency is superior to either metric alone for identifying high-quality reasoning data.

\begin{table}[t]
\centering
\small
\begin{adjustbox}{max width=\textwidth}
\begin{tabular}{lccc|ccc|ccc|c}
\toprule
\textbf{Dataset} & \multicolumn{3}{c}{\textbf{Up}} & \multicolumn{3}{c}{\textbf{Down}} & \multicolumn{3}{c}{\textbf{Non-reg.}} & \textbf{Acc} \\
\midrule
& {Prec} & {Rec} & {F1} & {Prec} & {Rec} & {F1} & {Prec} & {Rec} & {F1} &  \\
\midrule
All data & 0.07 $\pm$ 0.001 & 0.31 $\pm$ 0.005 & 0.12 $\pm$ 0.004 & 0.08 $\pm$ 0.003 & 0.64 $\pm$ 0.009 & 0.14 $\pm$ 0.003 & \textbf{0.93} $\pm$ 0.001 & \textbf{0.19} $\pm$ 0.002 & \textbf{0.31} $\pm$ 0.002 & 0.22 $\pm$ 0.003 \\
10\% random sampling & 0.06 $\pm$ 0.004 & 0.26 $\pm$ 0.007 & 0.10 $\pm$ 0.006 & 0.08 $\pm$ 0.005 & 0.63 $\pm$ 0.012 & 0.14 $\pm$ 0.004 & \textbf{0.93} $\pm$ 0.003 & \textbf{0.19} $\pm$ 0.005 & \textbf{0.31} $\pm$ 0.004 & 0.22 $\pm$ 0.006 \\
Keep lowest 10\% per class & \textbf{0.12} $\pm$ 0.005 & 0.36 $\pm$ 0.008 & \textbf{0.18} $\pm$ 0.007 & \textbf{0.11} $\pm$ 0.006 & 0.66 $\pm$ 0.014 & \textbf{0.19} $\pm$ 0.007 & 0.88 $\pm$ 0.004 & \textbf{0.19} $\pm$ 0.006 & \textbf{0.31} $\pm$ 0.005 & \textbf{0.25} $\pm$ 0.006 \\
Keep lowest 10\% global & \textbf{0.12} $\pm$ 0.006 & \textbf{0.38} $\pm$ 0.010 & \textbf{0.18} $\pm$ 0.008 & \textbf{0.11} $\pm$ 0.007 & \textbf{0.71} $\pm$ 0.016 & \textbf{0.19} $\pm$ 0.009 & 0.88 $\pm$ 0.005 & 0.06 $\pm$ 0.007 & 0.12 $\pm$ 0.008 & 0.16 $\pm$ 0.009 \\

\bottomrule
\vspace{0.5 mm}
\end{tabular}
\end{adjustbox}
\caption{\textbf{Synthetic data quality under different filtering strategies.} Random sampling selects traces uniformly, global uncertainty filtering selects the lowest 10\% CoCoA overall, and per-class filtering selects the lowest 10\% CoCoA within each class. Metrics include per-class precision (Prec), recall (Rec), F1, and overall coverage (Acc).}
\label{tab:filtering-ablation}
\end{table}

\begin{table}[t]
\centering
\small
\begin{adjustbox}{max width=\textwidth}
\begin{tabular}{lccc|ccc|ccc|c}
\toprule
\textbf{Dataset} & \multicolumn{3}{c}{\textbf{Up}} & \multicolumn{3}{c}{\textbf{Down}} & \multicolumn{3}{c}{\textbf{Non-reg.}} & \textbf{Acc}  \\
\midrule
& {Prec} & {Rec} & {F1} & {Prec} & {Rec} & {F1} & {Prec} & {Rec} & {F1} &   \\
\midrule
All data & 0.07 $\pm$ 0.001 & 0.31 $\pm$ 0.005 & 0.12 $\pm$ 0.004 & 0.08 $\pm$ 0.003 & 0.64 $\pm$ 0.009 & 0.14 $\pm$ 0.003 & \textbf{0.93} $\pm$ 0.001 & 0.19 $\pm$ 0.002 & \textbf{0.31} $\pm$ 0.002 & 0.22 $\pm$ 0.003 \\
Keep lowest 10\% CoCoA & \textbf{0.12} $\pm$ 0.005 & \textbf{0.36} $\pm$ 0.008 & \textbf{0.18} $\pm$ 0.007 & 0.11 $\pm$ 0.006 & 0.66 $\pm$ 0.014 & 0.19 $\pm$ 0.007 & 0.88 $\pm$ 0.004 & 0.19 $\pm$ 0.006 & \textbf{0.31} $\pm$ 0.005 & \textbf{0.25} $\pm$ 0.006 \\
Keep lowest 10\% consistency & 0.11 $\pm$ 0.005 & 0.30 $\pm$ 0.008 & 0.16 $\pm$ 0.007 & \textbf{0.12} $\pm$ 0.006 & \textbf{0.69} $\pm$ 0.014 & \textbf{0.20} $\pm$ 0.007 & 0.89 $\pm$ 0.004 & \textbf{0.20} $\pm$ 0.006 & \textbf{0.31} $\pm$ 0.005 & 0.24 $\pm$ 0.006 \\
Keep lowest 10\% perplexity & 0.09 $\pm$ 0.004 & 0.34 $\pm$ 0.007 & 0.14 $\pm$ 0.006 & 0.08 $\pm$ 0.005 & 0.64 $\pm$ 0.012 & 0.15 $\pm$ 0.004 & 0.91 $\pm$ 0.003 & 0.18 $\pm$ 0.005 & \textbf{0.31} $\pm$ 0.004 & 0.23 $\pm$ 0.006 \\

\bottomrule
\vspace{0.5 mm}
\end{tabular}
\end{adjustbox}
\caption{\textbf{Synthetic data quality when selecting traces based on different uncertainty metrics.} Perplexity measures fluency, consistency captures agreement across multiple traces, and CoCoA combines both signals. Metrics include per-class precision (Prec), recall (Rec), F1, and overall coverage (Acc).}
\label{tab:uncertainty-ablation}
\end{table}

%%%%%%%%%%%%%%%%%%%%%%%%%%%%%%%%%%%%%%%%%%%%%%%%
\section{Prompt} \label{app:prompts}

Figure~\ref{prompt2} shows the prompt used both for synthetic data generation, SFT training and evaluation.

\begin{figure}

\begin{tcolorbox}[colback=yellow!10!white, colframe=yellow!75!black, title=Prompt] \textbf{System message:}
You are an molecular and cellular biology expert analyzing gene regulation upon CRISPRi knockdown. First, provide your reasoning process within <think> </think> tags. Consider relevant pathways (e.g., cell-type specific biology, ribosome biogenesis, transcription, mitochondrial function, stress response), gene interactions, and cell-specific context. Then, choose one option from the following and place your choice within <answer> </answer> tags: 'upregulated', 'downregulated', or 'not differentially expressed'. Example: <think> [Your reasoning here] </think><answer> [upregulated / downregulated / not differentially expressed] </answer>

\textbf{User message:}
Analyze the regulatory effect of knocking down the {perturbation} gene on the {gene} gene in a single-cell {cell\_type} cell line using CRISPR interference.
\end{tcolorbox}
\caption{\textbf{Prompt template used for data generation, SFT, and evaluation.}}
\label{prompt2}
\end{figure}

%%%%%%%%%%%%%%%%%%%%%%%%

%\section{Computational Efficiency Analysis}
%\label{app:efficiency}

%\paragraph{Trace Generation Cost.}  
%For each input $(c,g,p)$, we generate $k{+}1$ traces (one greedy + $k=8$ sampled). With an average of $\sim$220 tokens per trace, this yields $\sim$2{,}000 tokens per example. On an A100 GPU, generating 50k examples requires $\sim$95 GPU-hours using Gemini 2.5 Pro, parallelized across 8 GPUs. This is a one-time cost amortized across all experiments.  

%By comparison, SynthPert’s label-conditioned generation requires both synthetic trace generation and label checking. The latter depends on experimental outcomes and cannot be parallelized across GPUs, making the cost effectively dominated by human/biological supervision. In practice, one wet-lab label requires days of bench work; thus, even a $10\times$ compute overhead is negligible compared to the cost of experimental supervision.  

%\paragraph{Filtering Overhead.}  
%Filtering adds little overhead: computing self-consistency and perplexity requires only lightweight cross-encoder evaluations, which run in milliseconds per example and are negligible compared to generation.  

%Overall, generation dominates compute cost but remains tractable on commodity hardware, while filtering overhead is minimal. More importantly, both are negligible compared to the prohibitive cost of wet-lab labels, underscoring the practical efficiency of uncertainty-filtered synthetic reasoning.  

\section{Computational Efficiency}
\label{app:efficiency}

We measured and verified the compute requirements of our generation and filtering pipeline, reporting both per-example and dataset-level costs. Table~\ref{tab:efficiency} summarizes the results for 50k examples.

\begin{table}[h]
\centering
\begin{adjustbox}{max width=\textwidth}
\begin{tabular}{lccc}
\toprule
\textbf{Component} & \textbf{Operation} & \textbf{Cost per example} & \textbf{Total (50k)} \\
\midrule
Trace generation & 9 traces ($\sim$2k tokens) & $\sim$0.0019 GPUh & 95 GPUh \\
Self-consistency & 8 cross-encoder passes & $\sim$0.000088 GPUh & 4.4 GPUh \\
Perplexity & log-probs of greedy trace & $\approx$0 (free at generation) & $\approx$0 \\
Aggregation + ranking & CPU-side sort & negligible & negligible \\
\midrule
\textbf{Total (ours)} & generation + filtering & $\sim$0.0020 GPUh & 99.4 GPUh \\
\bottomrule
\end{tabular}
\end{adjustbox}
\caption{\textbf{Measured compute costs for generating and filtering 50k synthetic examples.} GPUh denotes GPU-hours. Estimates are based on runs with Gemini 2.5 Pro (trace generation) and a medium cross-encoder (filtering).}
\label{tab:efficiency}
\end{table}

Overall, generation dominates compute requirements at $\sim$95 GPUh, while filtering adds only $\sim$4–5\% overhead (driven entirely by the cross-encoder self-consistency checks). Perplexity calculation incurs no additional cost when log-probs are retained during decoding. Aggregation and ranking are negligible. 

By comparison, SynthPert’s label-conditioned generation requires both synthetic trace generation and label checking. The latter depends on experimental outcomes and cannot be parallelized across GPUs, making the cost effectively dominated by human/biological supervision. In practice, one wet-lab label requires days of bench work; thus, even a $10\times$ compute overhead is negligible compared to the cost of experimental supervision.

\end{document}